\documentclass{article}

\usepackage{makeidx}
\usepackage{amsthm,amsmath,amssymb,amsfonts}
\usepackage{setspace,graphicx}
\usepackage{Generic}  
\usepackage[sort,longnamesfirst]{natbib}
\usepackage{url} 
\usepackage{xcolor}
\usepackage{bbm}
\usepackage{hyperref}
\usepackage{comment}
\usepackage{caption}
\usepackage{algorithm}
\usepackage{algpseudocode}
\usepackage[symbol]{footmisc}

\numberwithin{equation}{section}
\theoremstyle{plain}

\usepackage{bm}

\newcommand{\x}{\bm{x}}

\newcommand{\y}{\mathbf{y}}
\newcommand{\z}{\bm{z}}

\newcommand{\s}{\mathbf{s}}

\newcommand{\mparam}{\bm{\theta}}	
\newcommand{\vparam}{\bm{\phi}}	
\newcommand{\weight}{\mathbf{W}} 
\newcommand{\latent}{\bm{z}} 
\newcommand{\data}{\mathcal{D}} 

\newcommand{\wc}[1]{\textcolor{blue}{#1}}

\title{\huge\textbf{Bayesian Computation in Deep Learning}}
\date{}

\begin{document}
\maketitle
\doublespacing
%



\begin{center} \label{sec: label}
\begin{large}
{\em Wenlong Chen \footnote[1]{\label{footnote: eq}Equal Contribution}\footnote[2]{\label{footnote: ic}Imperial College London}, Bolian Li \footref{footnote: eq}\footnote[5]{\label{footnote: pu}Purdue University}, Ruqi Zhang \footnote[4]{\label{footnote: seq}Equal Contribution}\footref{footnote: pu} and Yingzhen Li \footref{footnote: seq}\footref{footnote: ic}}
\end{large}
\end{center}

\section{Introduction}
Bayesian computation has achieved profound success in many modeling tasks with statistics tools such as generalized linear models \citep{nelder1972generalized,dobson2018introduction}. Yet these traditional tools fail to produce satisfactory predictions for high-dimensional and highly complex data such as images, speech and videos. Deep Learning \citep{lecun2015deep} provides an attractive solution. At the time of late 2023, deep neural networks achieve accurate predictions for image classification \citep{dehghani2023scaling}, segmentation \citep{kirillov2023segment} and speech recognition tasks \citep{zhang2023google}. Meanwhile they have also demonstrated an astonishing capability for generating photo-realistic and/or artistic images \citep{rombach2022high}, music \citep{agostinelli2023musiclm} and videos \citep{liang2022nuwa}. Nowadays deep neural networks have become a standard modeling tool for many of the applications in AI and related fields, and the success of deep learning so far are based on training deterministic deep neural networks on big data. So one might ask: is there a place for Bayesian computation in modern deep learning? 

The answer is affirmative. Regarding prediction tasks that deep neural networks are primarily used for, Bayesian model averaging is capable for providing further improvements regarding accuracy \citep{wilson2020bayesian}. More importantly, quantifying uncertainty in predictions is crucial for increasing adoptions of neural networks in safety-critical domains such as healthcare \citep{gal2016thesis}. Decision-making tasks based on neural network predictions also require reliable uncertainty estimates \citep{savage1954bayes,jaynes2003probability}. All of these motivate the developments of Bayesian neural networks \citep{mackay:practical1992,neal1996bayesian,blundell2015bbp}, which applies Bayesian modeling to addressing the challenge of uncertainty quantification in deep learning.  
On the other hand, many popular neural network-based generative models for image/speech/video generation are based on deep latent variable models \citep{kingma2014auto,rezende:vae2014}, whose training requires inferring the unobserved latent variables. In this case posterior inference is a required step in generative model training, sharing many advantages and challenges as Bayesian computation in statistical modeling.

There are three main challenges for Bayesian computation in deep learning. First, the random variables to be inferred can be ultra high-dimensional, e.g., Bayesian neural networks require posterior inference over network weights which can be millions (if not billions) of variables. Second, the likelihood function is highly non-linear as a function of the random variable. In deep latent variable models the likelihood function is defined using a neural network, so that multiple distinct values of a latent variable can be mapped to e.g., the same mean of a Gaussian likelihood, which results in a highly multi-modal posterior. Third, the training data sizes for deep learning tasks are massive, e.g., in millions \citep{deng2009imagenet} to billions \citep{schuhmann2022laion}, so evaluating the likelihood functions on the entire dataset becomes challenging.
Traditional MCMC tools such as Metropolis-Hastings \citep{metropolis1953equation} and Gibbs sampling \citep{gelfand2000gibbs} fail to address these challenges in both fast and memory-efficient ways. These three challenges motivate the developments of \emph{approximate (Bayesian) inference} tools such as variational inference \citep{jordan1999vi,beal:vi2003,li2018approx,zhang2018advances}, stochastic-gradient MCMC \citep{welling2011bayesian,ma2015complete,zhangcyclical} and their advanced extensions.

This chapter provides an introduction to approximate inference techniques as Bayesian computation methods applied to deep learning models. We organize the chapter by presenting popular computational methods for (1) Bayesian neural networks and (2) deep generative models, explaining their unique challenges in posterior inference as well as the solutions. More specifically, the sections within this chapter are organised as follows.
\begin{itemize}
    \item Section \ref{sec:bnn} introduces Bayesian neural networks (BNNs), with a particular focus on the Bayesian computation tools applied to them. These computational tools are categorised into (1) Markov Chain Monte Carlo (MCMC, Section \ref{sec:bnn_sgmcmc}) and (2) Variational Inference (VI, Section \ref{sec:bnn_vi}). As we shall see, a common technique shared across the two themes is the use of stochastic gradient-based optimization \citep{bottou:ol1998}, which enables scaling of BNN computations to millions of datapoints.

    \item Section \ref{sec:dgm} discusses a variety of deep generative models which are used for density estimation and/or generating new samples from the data distribution. Specifically, MCMC serves as the key computation tool for sampling and training energy-based, score-based and diffusion models (Sections \ref{sec:dgm_ebm} \& \ref{sec:dgm_ddpm}), while VI is often employed in training deep latent variable models (Section \ref{sec:dgm_lvm}).
\end{itemize}

\vspace{-1em}
\section{Bayesian Neural Networks}
\label{sec:bnn}
Deep neural networks (DNNs), thanks to their expressiveness in modeling complex functions at scale \citep{lecun2015deep}, have gained significant attention in a wide range of applications, e.g., recognistion tasks in vision and speech domains \citep{dehghani2023scaling,kirillov2023segment,zhang2023google}.
In such supervised learning settings, a training dataset of $N$ input-output pairs $\data=\{(\bm{x}_n,\bm{y}_n)\}_{n=1}^N$ with $\bm{x} \in \mathbb{R}^{d_{in}}$ is provided.
We consider fitting an $L$-layer, width $d_h$, feed-forward DNN $f_{\mparam}(\bm{x}): \mathbb{R}^{d_{in}} \rightarrow \mathbb{R}^{d_{out}}$ with weight and bias parameters $\mparam=\{\weight^l, \mathbf{b}^l\}_{l=1}^L$, whose functional form is
\begin{equation}
f_{\mparam}(\bm{x})=\weight^L g(\weight^{L-1}g(\cdot\cdot\cdot g(\weight^1 \bm{x}+\mathbf{b}^1))+\mathbf{b}^{L-1})+\mathbf{b}^L,
\label{eq:feed_forward_dnn}
\end{equation}
\[\weight^1 \in \mathbf{R}^{d_h \times d_{in}}, \mathbf{b}^1 \in \mathbf{R}^{d_{in}}, \weight^l \in \mathbf{R}^{d_h \times d_h}, \mathbf{b}^l \in \mathbf{R}^{d_h}, l = 2, ..., L-1, \weight^L \in \mathbf{R}^{d_{out} \times d_h}, \mathbf{b}^L \in \mathbf{R}^{d_{out}}. \]
Here a non-linear activation $g(\cdot)$ (e.g., ReLU \citep{hinton2010relu}) is used to make $f_{\mparam}(\bm{x})$ a non-linear function w.r.t.~$\bm{x}$.
For regression problems, we use the neural network to produce the prediction $\hat{\bm{y}} = f_{\mparam}(\bm{x})$. In such case $d_{out}$ equals to the dimensionality of $\bm{y}$, and the likelihood of the model is typically defined as a Gaussian distribution with a hyper-parameter $\sigma$:
\begin{equation}
    p(\bm{y} | \bm{x}, \mparam) = \mathcal{N}(\bm{y}; f_{\mparam}(\bm{x}), \sigma^2 \mathbf{I}).
\end{equation}
On the other hand, for classification problems, categorical likelihood is used, and the neural network returns the logit of the probability vector of the categorical distribution: assuming $\bm{y} \in \{1, ..., C \}$, then $d_{out} = C$ and
\begin{equation}
    [p(\bm{y} = 1| \bm{x}, \mparam), ..., p(\bm{y} = C| \bm{x}, \mparam)] = \text{softmax}(f_{\mparam}(\bm{x})), \quad \text{softmax}(\bm{l}) := \frac{\exp(\bm{l})}{ \sum_{c=1}^C \exp(\bm{l}_c)}.
\end{equation}
The power of DNNs lies in their capability in learning expressive feature representations for the input data. To see this, we rewrite the neural network feed-forward pass up to the $L-1^{\text{th}}$ layer in Eq.~\eqref{eq:feed_forward_dnn} as
$
\Phi_{\mparam}(\bm{x})=\weight^{L-1} g(\weight^{L-2}g(\cdot\cdot\cdot g(\weight^1 \bm{x}+\mathbf{b}^1))+\mathbf{b}^{L-2})+\mathbf{b}^{L-1}.
$
Then the neural network $f_{\mparam}(\bm{x}) = \weight^L \Phi_{\mparam}(\bm{x}) + \mathbf{b}^L$ and the associated likelihood model can be viewed as a (generalized) linear regression model with non-linear features $\Phi_{\mparam}(\bm{x})$. Different from generalized linear regression models, however, these non-linear features are also learned jointly with the last-layer parameters $\{ \weight^L, \mathbf{b}^L \}$. While this joint learning procedure returns expressive features and improved predictive accuracy, the weight and bias parameters $\{ \weight^l, \mathbf{b}^l \}_{l=1}^L$ are often less interpretable in any statistical sense, although sparsity methods may provide explanations akin to feature selection \citep{lemhadri2021lassonet,ghosh2019model}. See later paragraphs for a further discussion in Bayesian computation context.

A typical estimation procedure for the parameters $\mparam$ is maximum likelihood estimation (MLE), which finds the optimal neural network parameters as
\[\mparam^* = \arg\max_{\mparam} \ \sum_{n=1}^N \log p(\bm{y}_n | \bm{x}_n, \mparam). \]
However, when the dataset size is substantially smaller than the network parameter size, MLE estimates may overfit. Maximum a Posteriori (MAP) solutions can help alleviate the overfitting issue, however, both MLE and MAP provide point estimates only for the network parameters $\mparam$, leaving the underlying uncertainty uncaptured.

\begin{figure}[t]
    \centering
\includegraphics[width=0.47\textwidth]{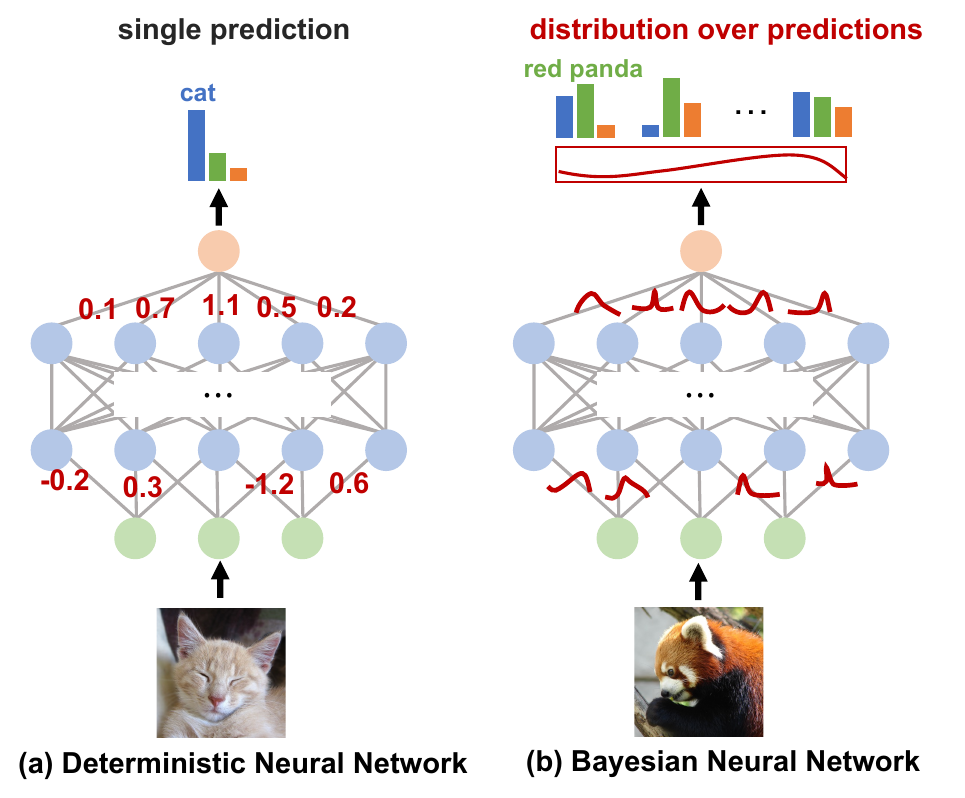}
    \caption{Difference between standard deep neural network and Bayesian neural network.}
    \label{fig: bnn}
    \vspace{-1em}
\end{figure}

Bayesian inference is a principled framework to obtain reliable uncertainty in DNNs \citep{gal2016thesis}. Instead of using a point estimate of $\mparam$, Bayesian inference computes a posterior distribution over $\mparam$ to explain the observations with uncertainty estimates (See Figure \ref{fig: bnn}): 
\begin{equation}
    p(\mparam|\data)=\frac{p(\mparam)p(\data|\mparam)}{p(\data)}=\frac{p(\mparam)p(\data|\mparam)}{\int p(\mparam)p(\data|\mparam)d\mparam}.
\label{eq:bayespost}
\end{equation}
Here $p(\mparam)$ is the prior distribution representing our belief about the model parameters without observing any data points, and $p(\data|\mparam)$ is the likelihood. Often we assume i.i.d.~likelihood, which means
$p(\data|\mparam) := \prod_{n=1}^N p(\bm{y}_n | \bm{x}_n, \mparam)$. 
In the rest of the chapter we also write $p(\bm{y} | f_{\mparam}(\bm{x})) := p(\bm{y} | \bm{x}, \mparam)$ since the distributional parameters of the likelihood mainly depends on the output of the neural network.
The resulting neural networks are called Bayesian neural networks (BNNs) \citep{neal1996bayesian,mackay:practical1992,hinton:mdl1993}. For predictions, the uncertainty in $\mparam$ translates to the predictive uncertainty for $(\bm{x}^\ast, \bm{y}^\ast)$, via the following \emph{posterior predictive distribution}:
\begin{equation}
    p(\bm{y}^\ast|\bm{x}^\ast,\data) = \int p(\bm{y}^\ast|\bm{x}^\ast,\mparam) p(\mparam|\data) d\mparam.
\label{eq:bayespred}
\end{equation}

With the posterior predictive distribution (Eq.~\eqref{eq:bayespred}) one can compute uncertainty measures from it. In regression problems a typical uncertainty measure is the variance of the prediction $V[\bm{y}^* | \bm{x}^*] = E_{p(\bm{y}^\ast|\bm{x}^\ast,\data)}[(\bm{y}^* - \bm{\mu}^*)(\bm{y}^* - \bm{\mu}^*)^{\top}]$ with $\bm{\mu}^* = E_{p(\bm{y}^\ast|\bm{x}^\ast,\data)}[\bm{y}^*]$ as the predictive posterior mean. In classification where the likelihood $p(\bm{y} | \bm{x}, \mparam)$ is typically categorical, one can also use the entropy $H[p(\bm{y}^\ast|\bm{x}^\ast,\data)] = -\sum_{c=1}^C p(\bm{y}^\ast = c |\bm{x}^\ast,\data) \log p(\bm{y}^\ast = c |\bm{x}^\ast,\data)$ to express uncertainty. These two quantities represent \emph{total} uncertainty, which can be further decomposed into \emph{epistemic} and \emph{aleatoric} uncertainty measures, for example \citep{houlsby2011bayesian,gal2016thesis}:
\begin{equation}
    \underset{(\text{total uncertainty})}{H[p(\bm{y}^\ast|\bm{x}^\ast,\data)]} = \underset{(\text{aleatoric uncertainty})}{E_{p(\mparam | \data)}[H[p(\bm{y}^\ast|\bm{x}^\ast,\mparam)]]} + \underset{(\text{epistemic uncertainty})}{I[\bm{y}^*; \mparam | \bm{x}^*, \data]}.
\label{eq:total_uncertainty}
\end{equation}
The conditional entropy $E_{p(\mparam | \data)}[H[p(\bm{y}^\ast|\bm{x}^\ast,\mparam)]] = E_{p(\mparam | \data)} E_{p(\bm{y}^\ast|\bm{x}^\ast,\mparam)}[-\log p(\bm{y}^\ast|\bm{x}^\ast,\mparam)]$ reveals the average ``variability'' (under posterior $p(\mparam | \data)$) of each neural network's predictive distribution $p(\bm{y}^* | \bm{x}^*, \mparam)$, and it aims to quantify the intrinsic randomness in data due to e.g.,~noisy measurements.
On the other hand, there are two explanations for epistemic uncertainty, originated from two equivalent definitions of mutual information. 
The first definition reads
\[I[\bm{y}^*; \mparam | \bm{x}^*, \data] = \mathbb{E}_{p(\bm{y}^* | \bm{x}^*, \data)} [\mathrm{KL}[ p(\mparam | \bm{y}^*, \bm{x}^*, \data) || p(\mparam | \data)]], \]
which quantifies the expected information gain for $\mparam$ (i.e., reduction of uncertainty in $p(\mparam | \data)$) when adding the prediction $\bm{y}^*$ on $\bm{x}^*$ to the observations. This concept of ``epistemic uncertainty reduction'' is particularly useful in data acquisition tasks following Bayesian principles \citep{houlsby2011bayesian}. The second definition of mutual information comes from re-arranging terms in the first definition:
\[I[\bm{y}^*; \mparam | \bm{x}^*, \data] = \mathbb{E}_{p(\mparam | \data)} [\mathrm{KL}[ p(\bm{y}^* | \bm{x}^*, \mparam) || p(\bm{y}^* | \bm{x}^*, \data)]]. \]
It measures the average deviation (under the posterior $p(\mparam | \data)$) of a single neural network's prediction $p(\bm{y}^* | \bm{x}^*, \mparam)$ from the posterior predictive $p(\bm{y}^* | \bm{x}^*, \data)$. Therefore a high value of mutual information means the neural networks parameterized by different $\mparam$ from high density regions of $p(\mparam | \data)$ have large disagreement in their prediction results, and the BNN is said to be epistemically uncertain about its predictions.

Unfortunately, Eq.~\eqref{eq:bayespost} is intractable since the marginal likelihood requires a complex integral $p(\data)=\int p(\mparam)p(\data|\mparam)d\mparam$, which has no analytic solution due to non-linearities in DNNs. Furthermore, numerical integration is infeasible due to extremely high-dimensional $\mparam$, usually over million or even billions in modern DNNs \citep{krizhevsky2012imagenet,dehghani2023scaling}. Therefore, we have to resort to approximate inference techniques \citep{li2018approx}, which uses an approximate posterior distribution $q(\mparam) \approx p(\mparam | \data)$ to replace the exact posterior in Bayesian inference tasks. In such case posterior predictive distribution (Eq.~\eqref{eq:bayespred}) is approximated by (typically also with Monte Carlo estimate):
\begin{equation}
    p(\bm{y}^* | \bm{x}^*, \data) \approx \int p(\bm{y}^\ast|\bm{x}^\ast,\mparam) q(\mparam) d\mparam \approx \frac{1}{M} \sum_{m=1}^M p(\bm{y}^\ast|\bm{x}^\ast,\mparam_m), \quad \mparam_m \sim q(\mparam).
\label{eq:approximate_predictive_distribution}
\end{equation}
This approximate predictive posterior can then be used subsequently for computing the uncertainty measures such as predictive variance or entropy in Eq.~\eqref{eq:total_uncertainty}.
We will discuss two classes of approximations: (1) stochastic gradient MCMC \citep{welling2011bayesian,chen2014stochastic}, which constructs $q(\mparam) = \frac{1}{M} \sum_{m=1}^M \delta(\mparam = \mparam_m)$ using samples $\{ \mparam_{m} \}_{m=1}^N$ obtained via an MCMC algorithm, and (2) variational inference \citep{beal:vi2003,jordan1999vi} where $q(\mparam)$ often has a parametric form and is obtained via an optimization procedure. 

Before proceeding to the technical details, an interesting note is that for BNNs, the quality of posterior approximation $q(\mparam) \approx p(\theta | \data)$ is mostly evaluated via the corresponding approximate posterior predictive  (Eq.~\eqref{eq:approximate_predictive_distribution}) rather than by direct inspection. First, symmetry exists in neural network parameters $\mparam$ (i.e., there exists $\mparam_1 \neq \mparam_2$ such that $f_{\mparam_1} = f_{\mparam_2}$, e.g., by swaping two hidden units of a hidden layer in the neural network of $f_{\mparam_1}$ and denoting the corresponding neural network's weight by $\mparam_2$ \citep{phuong2020functional,rolnick2020reverse}), meaning that the exact posterior $p(\mparam | \data)$ is highly multi-modal. Second and most distinctively from traditional statistical models, due to the emphasis of prediction accuracy for deep learning models, the network weights $\mparam$ are not designed to have an associated physical meaning, and in most cases even the exact posterior $p(\mparam | \data)$ is not interpretable by humans. Instead, the posterior predictive distribution (Eq.~\eqref{eq:approximate_predictive_distribution}) directly expresses the uncertainty in prediction, which can then be analyzed and used in further decision-making tasks based on the predictions.


\begin{algorithm}[t]
\caption{SG-MCMC methods for sampling from BNN posteriors}\label{alg:sgmcmc_bnn}
\begin{algorithmic}[1]
\Require Dataset $\data = \{ (\x_n, \y_n) \}_{n=1}^N$, initialized parameter $\mparam_1$,  batch-size $B$, SG-MCMC step sizes $\{\alpha_t \}$, total steps $T$.
\For{$t = 1, ..., T$}
\State Sample a mini-batch of datapoints $\bm{\Xi} = \{ (\x, \y) \} \sim \data^B$ with $| \bm{\Xi} | = B$
\State Get the stochastic gradient $\nabla_{\mparam_t}\widetilde{U}(\mparam_t)$ via automatic differentiation
\State Obtain $\mparam_{t+1}$ through an SG-MCMC update, using the stochastic gradient $\nabla_{\mparam_t}\widetilde{U}(\mparam_t)$ and the step size $\alpha_t$
\EndFor
\end{algorithmic}
\end{algorithm}

\subsection{Stochastic Gradient MCMC}
\label{sec:bnn_sgmcmc}
Classical MCMC methods require computation over the entire dataset to obtain the true likelihood or its gradient, making them prohibitively expensive for large datasets. For example, the standard Metropolis-Hastings (MH) algorithm is unsuitable because the accept/reject step necessitates computing the likelihood, which involves summing across the entire dataset. To mitigate this computational burden, minibatched MH algorithms have been developed~\cite{korattikara2014austerity,maclaurin2014firefly,quiroz2018speeding,zhang2020asymptotically}. These algorithms utilize a subset of the data in the MH step to reduce costs. However, they often rely on strong assumptions, such as bounded log-likelihood or bounded gradients, which render them impractical for deep neural networks. 
Despite MCMC methods, notably Hamiltonian Monte Carlo~\citep{duane1987hybrid,neal2010mcmc}, being regarded as the gold standard for Bayesian computations, they are rarely applied in modern BNNs due to their large computational cost. To address this challenge, Stochastic Gradient MCMC (SG-MCMC) methods~\citep{welling2011bayesian,chen2014stochastic,ma2015complete,zhangcyclical} have been proposed. These methods introduce stochastic gradients into MCMC sampling steps, relying only on a subset of data at each iteration. Consequently, they demand significantly less memory compared to standard MCMC approaches. SG-MCMC enables the use of sampling methods in large-scale Bayesian Neural Networks, such as ResNets~\citep{he:resnet2016}. In this section, we introduce three popular variants of SG-MCMC algorithms in deep learning. A summary of the procedure of SG-MCMC algorithms is provided in Algorithm~\ref{alg:sgmcmc_bnn}.

\vspace{-1em}
\subsubsection{Stochastic Gradient Langevin Dynamics}
Stochastic Gradient Langevin Dynamics (SGLD)~\citep{welling2011bayesian} stands as the first SG-MCMC algorithm. In contrast to the full energy function $U(\mparam)=-\sum_{(\bm{x}, \bm{y}) \in\data}\log{p(\bm{x}|\mparam)}-\log{p(\mparam)}$ for the exact posterior distribution $p(\mparam|\data)\propto\exp(-U(\mparam))$, the SGLD algorithm considers a mini-batch energy function with a mini-batch of datapoints $\bm{\Xi}\subseteq\data$: 
\begin{equation}
    \widetilde{U}(\mparam)=-\frac{|\data|}{|\bm{\Xi}|}\sum_{(\bm{x}, \bm{y})\in\bm{\Xi}}\log{p(\bm{y} | \bm{x}, \mparam)}-\log{p(\mparam)},
\end{equation}
By resampling the minibatch $\bm{\Xi}$ at each iteration, the update rule of SGLD follows the \emph{underdamped Langevin dynamics}, replacing the true gradient with the stochastic gradient:
\begin{equation}
    \mparam_{t+1} \gets \mparam_t-\alpha_t\nabla_{\mparam_t}\widetilde{U}(\mparam_t)+\sqrt{2\alpha_t}\cdot\bm{\epsilon}_t,
\end{equation}
where $\alpha_t$ is the step size at iteration $t$ and $\bm{\epsilon}_t\sim \mathcal{N}(\bm{0},\bm{I})$ is the standard Gaussian noise. Compared with the update rule of stochastic gradient descent (SGD), the only difference is the addition of the appropriate Gaussian noise.
\citet{welling2011bayesian} shows that when the step size $\alpha_t$
is sufficiently small, SGLD becomes standard unadjusted Langevin dynamics~\citep{roberts1996exponential}, as the Gaussian noise term dominates over the stochastic noise. To ensure a low asymptotic estimation error, a decaying step size schedule is typically applied \citep{bottou:ol1998,welling2011bayesian,ma2015complete}, satisfying (i) $\sum_{t=1}^\infty\alpha_t=\infty$ and (ii) $\sum_{t=1}^\infty\alpha_t^2<\infty$. The first requirement guarantees that the sampler will reach the high probability areas regardless of initialization, and the second requirement ensures a small asymptotic error~\citep{welling2011bayesian}.

\begin{figure}[t]
    \centering
\includegraphics[width=0.5\textwidth]{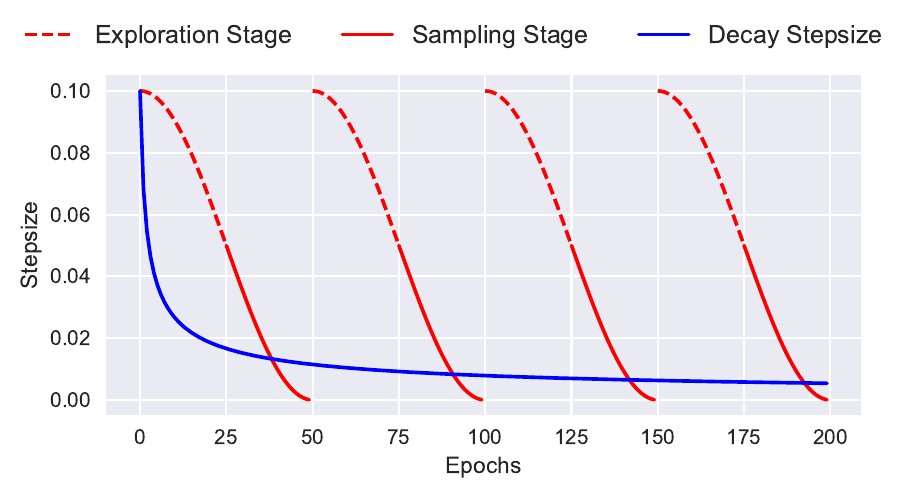}
    \caption{Comparison between the cyclical stepsize schedule (red) and the traditional decreasing stepsize schedule (blue) for SG-MCMC algorithms. Adapted from
\citet{zhangcyclical}. Used with the kind permission of Ruqi Zhang.}
    \label{fig: csgmcmc}
\end{figure}
\subsubsection{Cyclical Stochastic Gradient MCMC}
As mentioned in the last section, decreasing step sizes for SG-MCMC is typically required to ensure asymptotic accuracy of posterior estimation. However, small step sizes will significantly reduce the sampler's ability to explore the entire domain and to exploit each mode in detail efficiently, resulting in large approximation errors in finite time~\citep{vollmer2016exploration,chen2015convergence}. This issue is exacerbated when estimating Bayesian neural network posteriors, which are complicated and highly multi-modal~\citep{neal1996bayesian,izmailov2021bayesian}. 

To solve this problem, \cite{zhangcyclical} proposes a cyclical step size schedule for SG-MCMC, depicted in Figure~\ref{fig: csgmcmc}. This schedule defines the step size at iteration $t$ as
\begin{equation}
\alpha_t = \frac{\alpha_0}{2}
\left[\cos\left(\frac{\pi~\text{mod}(t-1,\lceil T/M\rceil)}{\lceil T/M\rceil}\right)+1
\right], \label{eq:cyclic_lr}
\end{equation}
where $\alpha_0$ is the initial step size, $M$ is the number of cycles and $T$ is the number of total iterations. The cyclical step size schedule enables the sampler to efficiently explore various modes and accurately characterize local modes at a fine scale. 

To efficiently sample from the multi-modal posteriors in Bayesian neural networks, other strategies include SG-MCMC methods combined with parallel tempering~\citep{deng2020non} and flat histograms~\citep{deng2020contour}.

\subsubsection{Stochastic Gradient Hamiltonian Monte Carlo}
Hamiltonian Monte Carlo (HMC) has achieved the gold standard of performance with small neural networks [Here, refer to the Chapter of HMC]. It incorporates a kinetic energy term, characterized by a set of ``momentum" auxiliary variables. The naive stochastic gradient HMC adopts a mini-batch Hamiltonian function:
\begin{equation}
    \widetilde{H}(\mparam,\bm{r})=\widetilde{U}(\mparam)+\frac{1}{2}\bm{r}^T\bm{M}^{-1}\bm{r}=-\frac{|\data|}{|\bm{\Xi}|}\sum_{(\bm{x}, \bm{y}) \in\bm{\Xi}}\log{p(\bm{y}|\bm{x}, \mparam)}-\log{p(\mparam)}+\frac{1}{2}\bm{r}^T\bm{M}^{-1}\bm{r},
\end{equation}
where $\bm{r}$ is the momentum vector and $\bm{M}$ is a positive definite mass matrix. 
Despite its intuitive appeal, \citet{chen2014stochastic} indicates that the naive adaption of HMC algorithm to the stochastic gradient version can diverge from the target distribution. This is because the joint distribution of parameters and momentum (i.e., $\pi(\mparam,\bm{r})\propto\exp(-\widetilde{H}(\mparam,\bm{r}))$) are not invariant for this Markov chain due to the introduction of stochastic noise. For corrections, \citet{chen2014stochastic} introduces a ``friction" term $C\bm{M}^{-1}\bm{r}$ where $C$ is the friction weight, to the momentum updating:
\begin{equation}
    \frac{d\mparam}{dt}=\bm{M}^{-1}\bm{r}~~~~\text{and}~~~~\frac{d\bm{r}}{dt}=-\nabla_{\mparam}U(\mparam)-C\bm{M}^{-1}\bm{r}+\sqrt{2C}\bm{\epsilon}, \ \bm{\epsilon} \sim \mathcal{N}(\bm{0},\bm{I}).
\end{equation}
This friction term helps decrease the Hamiltonian $\widetilde{H}(\mparam,\bm{r})$, which counters the effects of random noise. The modified dynamics is commonly denoted as \emph{second-order Langevin dynamics}~\citep{wang1945theory}.


\vspace{-3em}
\subsection{Variational Inference}
\label{sec:bnn_vi}
Variational inference (VI) \citep{jordan1999vi,beal:vi2003} is a class of approximate inference techniques, which approximates the exact posterior using a tractable distribution, typically from a parametric distribution family $\mathcal{Q} := \{q_{\vparam}(\mparam) \}$ such as factorized Gaussians. The best approximate posterior $q_{\vparam^*}(\mparam)$, which is used as $q(\mparam)$ in Eq.~\eqref{eq:approximate_predictive_distribution} and also named as the optimal variational distribution, is selected to be the one that is the closest to the posterior $p(\mparam|\data)$ within the parametric family as measured by a divergence measure between distributions. 
The Kullback-Leibler (KL) divergence \citep{kullback:divergence1951} is a popular choice for VI: here the approximate posterior is obtained by minimizing the KL-divergence with respect to the parameters $\vparam$ of the chosen parametric distributions family:
\begin{equation}
    \vparam^* = \arg\min_{q_{\vparam} \in \mathcal{Q}} KL[q_{\vparam}(\mparam)\|p(\mparam|\data)], \quad KL[q_{\vparam}(\mparam)\|p(\mparam|\data)] = \int q_{\vparam}(\mparam) \log \frac{q_{\vparam}(\mparam)}{p(\mparam|\data)} d\mparam.
\label{eq:kl}
\end{equation}
Again $p(\mparam|\data)$ is intractable due to the intractable marginal likelihood $p(\data)$, meaning that direct minimization of the KL-divergence is intractable. Fortunately, this KL-divergence minimization task is equivalent to maximizing the following Evidence Lower Bound (ELBO), which is obtained by subtracting the KL-divergence from the log marginal-likelihood which is constant w.r.t.~the variational parameters $\vparam$:
\begin{equation}
    \begin{split}
        \mathcal{L}_{ELBO}(\vparam) &:=\log p(\data) -KL[q_{\vparam}(\mparam)\|p(\mparam|\data)]\\
        &=\underset{(i)}{E_{q_{\vparam}(\mparam)}[\log p(\data|\mparam)]} - \underset{(ii)}{KL[q_{\vparam}(\mparam)\|p(\mparam)]}.
    \end{split}
\label{eq:elbo}
\end{equation}
In Eq. \eqref{eq:elbo}, the ELBO is decomposed into (i) the expectation of log-likelihood under $q_{\vparam}$, optimization of which encourages better data fitting, and (ii) a KL regularizer, optimization of which encourages $q_{\vparam}$ to be close to the prior. Combining both terms, the ELBO optimization encourages $q_{\vparam}$ to find small but meaningful deviations from the prior that can explain the observed data.
In practice, the expected log-likelihood term (i) in the ELBO can not be analytically evaluated. Therefore an unbiased Monte Carlo estimate is applied to the ELBO with $M$ samples from the variational posterior \citep{graves2011practical,blundell2015bbp}:
\begin{equation}
    \hat{\mathcal{L}}_{ELBO}(\vparam) = \frac{1}{M} \sum_{m=1}^M \log p(\data|\mparam_m) - KL[q_{\vparam}(\mparam)\|p(\mparam)], \qquad \mparam_m \stackrel{\text{iid}}{\sim} q_{\vparam}(\mparam) .
\label{eq:elbo_mc_estimate}
\end{equation}
The KL regularizer term in ELBO may be evaluated analytically for some choices of prior and variational family (e.g., when they are both Gaussians). When analytic solution is not available, this KL term can also be estimated with Monte Carlo.

VI will return the exact posterior if $p(\mparam |\data) \in \mathcal{Q}$, which is not the case in practice. Instead, due to limited computational budget one would prefer simple $q$ distributions that enjoy fast computations and low memory costs. In BNN context, one of the most popular choices is mean-field Gaussian approximation (i.e., factorized Gaussian) \citep{blundell2015bbp}:
\begin{equation}
    q_{\vparam}(\mparam) = \prod_{l=1}^L q_{\vparam}(\weight^l)q_{\vparam}(\mathbf{b}^l), \ q_{\vparam}(\weight^l) = \prod_{ij} \mathcal{N}(\weight^l_{ij}; \mathbf{M}^l_{ij}, (\mathbf{S}^l_{ij})^2), \ q_{\vparam}(\mathbf{b}_l) = \prod_j \mathcal{N}(\mathbf{b}^l_{j}; \mathbf{m}^l_{j}, (\s^l_{j})^2).
\label{eq:mean_field_bnn_q_distribution}
\end{equation}
Here the variational parameters are $\vparam = \{\mathbf{M}^l, \mathbf{m}^l, \mathbf{S}^l, \mathbf{s}^l \}_{l=1}^L$ which are optimized by maximizing the ELBO, typically with Monte Carlo estimates (Eq.~\eqref{eq:elbo_mc_estimate}). Therefore, compared with deterministic DNNs which directly optimizes the weight parameters $\mparam=\{\weight^l, \mathbf{b}^l\}_{l=1}^L$, mean-field VI for BNNs doubles the amount of parameters to be optimized (hence twice amount of memory consumption), due to the additional variance parameters $\{\mathbf{S}^l, \mathbf{s}^l \}_{l=1}^L$. 

The optimization process of ELBO is conducted via gradient ascent, and in order to apply back-propagation with e.g., automatic differentiation techniques \citep{auto2018baydin}, the \emph{reparameterization trick} \citep{kingma2014auto} is introduced for a range of variational distribution families including Gaussians. This approach assumes the sampling operation $\mparam \sim q_{\vparam}(\mparam)$ is defined as transforming an auxiliary noise variable $\bm{\epsilon} \sim p_{base}(\bm{\epsilon})$ through a function $T_{\vparam}(\bm{\epsilon})$ that is differentiable w.r.t.~$\vparam$:
\begin{equation}
   \mparam \sim q_{\vparam}(\mparam) \quad \Leftrightarrow \quad \mparam = T_{\vparam}(\bm{\epsilon}), \ \bm{\epsilon} \sim p_{base}(\bm{\epsilon}).
\end{equation}
For mean-field Gaussians, e.g., $q_{\vparam}(\weight^l)$ in Eq.~\eqref{eq:mean_field_bnn_q_distribution}, this transformation is defined as $\weight^l = T_{\vparam}(\bm{E}^l) := \mathbf{M}^l + \sqrt{\mathbf{S}^l} \odot \bm{E}^l$, with $\bm{E}^l_{ij} \sim \mathcal{N}(0, 1)$, and $\sqrt{\cdot}$ and $\odot$ denote element-wise square-root and multiplication, respectively. This allows us to rewrite the Monte Carlo estimate of the expected log-likelihood term in Eq.~\eqref{eq:elbo} using change-of-variable rules, where we collect the mean and variance parameters into $\bm{\mu} := \{\mathbf{M}^l, \mathbf{m}^l \}_{l=1}^L$ and $\bm{\sigma} := \{\sqrt{\mathbf{S}^l}, \sqrt{\mathbf{s}^l} \}_{l=1}^L$:
\begin{equation}
    E_{q_{\vparam}(\mparam)}[\log p(\data|\mparam)] \approx \frac{1}{M} \sum_{m=1}^M \log p(\data|\bm{\mu} + \bm{\sigma} \odot \bm{\epsilon}_m), \quad \bm{\epsilon}_m \stackrel{\text{iid}}{\sim} \mathcal{N}(\bm{0}, \mathbf{I}).
\end{equation}
The corresponding gradient w.r.t.~$\mathbf{M}^l$ for example can then be computed via back-propagation. Recall that $p(\data|\mparam) := \prod_{n=1}^N p(\bm{y}_n | \bm{x}_n, \mparam)$ with $p(\bm{y} | \bm{x}, \mparam) := p(\bm{y} | f_{\mparam}(\bm{x}))$, then
$$\nabla_{\mathbf{M}^l} E_{q_{\vparam}(\mparam)}[\log p(\data|\mparam)] \approx \frac{1}{M} \sum_{m=1}^M \sum_{n=1}^N \nabla_{\mathbf{M}^l} f \nabla_f \log p(\bm{y}_n| f_{\bm{\mu} + \bm{\sigma} \odot \bm{\epsilon}_m}(\bm{x}_n)), \quad \bm{\epsilon}_m \stackrel{\text{iid}}{\sim} \mathcal{N}(\bm{0}, \mathbf{I}).$$
In practice, using $M=1$ sample, together with proper initialization of the $\vparam$ parameters, is sufficient for the gradient estimation.
Similar operations can be applied to computing the gradient of the KL term in the ELBO when Monte Carlo estimate is also required.

Lastly, evaluating $\log p(\data|\mparam)$ requires processing the entire dataset $\data$ which can be costly in big data setting. Similar to training deterministic DNNs and running SG-MCMC for BNNs, stochastic optimization methods such as stochastic gradient descent (SGD) \citep{bottou:ol1998} are also applicable to variational inference. With $\bm{\Xi}\subseteq\data$ a mini-batch sampled from the entire dataset, the ELBO can be estimated as (with $M=1$ Monte Carlo sample and the reparameterization trick):
\begin{equation}
    \tilde{\mathcal{L}}_{ELBO}(\vparam) = \frac{|\data|}{|\bm{\Xi}|} \sum_{(\bm{x}, \bm{y}) \in \bm{\Xi}} \log p(\bm{y}| f_{\bm{\mu} + \bm{\sigma} \odot \bm{\epsilon}}(\bm{x})) - KL[q_{\vparam}(\mparam)\|p(\mparam)], \quad \bm{\epsilon} \sim \mathcal{N}(\bm{0}, \mathbf{I}).
\label{eq:elbo_mc_estimate_minibatch}
\end{equation}
Automatic differentiation can be directly applied to $\tilde{\mathcal{L}}_{ELBO}(\vparam)$, and the optimization of this ELBO estimate has the same time complexity as running stochastic gradient descent training of deterministic DNNs. All the above techniques combined enable BNNs with mean-field Gaussian VI to scale up to modern DNN architectures such as ResNets \citep{he:resnet2016}, and a summary of the training procedure is provided in Algorithm \ref{alg:mfvi_bnn}.

\begin{algorithm}[t]
\caption{SGD Training for a BNN with mean-field Gaussian VI}\label{alg:mfvi_bnn}
\begin{algorithmic}[1]
\Require Dataset $\data = \{ (\x_n, \y_n) \}_{n=1}^N$, initialised variational parameters $\vparam = \{ \bm{\mu}, \bm{\sigma
} \}$,  batch-size $B$, SGD step sizes $\{\alpha_t \}$, SGD total steps $T$.
\For{$t = 1, ..., T$}
\State Compute the KL term in the ELBO:
$\tilde{\mathcal{L}}_{ELBO}(\vparam) \gets - KL[q_{\vparam}(\mparam)\|p(\mparam)]$ 
\State Sample BNN weights with the reparameterization trick: $\mparam \gets \bm{\mu} + \bm{\sigma} \odot \bm{\epsilon}$, $\bm{\epsilon} \sim \mathcal{N}(\bm{0}, \mathbf{I})$
\State Sample a mini-batch of datapoints $\bm{\Xi} = \{ (\x, \y) \} \sim \data^B$ with $| \bm{\Xi} | = B$
\For{$(\x, \y) \in \bm{\Xi}$}
    \State Add-in data likelihood: $\tilde{\mathcal{L}}_{ELBO}(\vparam) \gets \tilde{\mathcal{L}}_{ELBO}(\vparam) + \frac{|\data|}{|\bm{\Xi}|} \log p(\bm{y}| f_{\mparam}(\bm{x}))$
\EndFor
\State Compute SGD updates via automatic differentiation:
$\vparam \gets \vparam + \alpha_t \nabla_{\mparam} \tilde{\mathcal{L}}_{ELBO}(\vparam)$
\EndFor
\end{algorithmic}
\end{algorithm}

\subsubsection{Alternative Divergences}

Standard VI based on minimizing $KL[q_{\vparam}(\mparam)\|p(\mparam|\data)]$, while enjoying many computational advantages,  tends to underestimate posterior uncertainty \citep{turner:two_problems2011}. The Gaussian approximation from standard VI often captures only one mode of the exact posterior, which has been referred to as the mode-seeking phenomenon \citep{miguel2015alpha, li2016renyi}. This issue can be mitigated by extending the VI framework to other divergences. For example, $\alpha$-divergence is a subset of $f$-divergence family \citep{csiszar:divergence1963} with a hyper-parameter $\alpha$ determining the degree of the posterior mass coverage of the variational distribution \citep{renyi:divergence1961,van_erven:renyi2014}:

%
%
\begin{equation}
    D_{\alpha}[ p(\mparam|\data) \| q_{\vparam}(\mparam) ] = \frac{1}{(\alpha-1)} \log \int p(\mparam|\data)^{\alpha} q_{\vparam}(\mparam)^{1-\alpha} d\mparam.
    \label{eq:alpha_divergence}
\end{equation}
When $\alpha$ is set to approach $1$ and $0$, we recover the two special cases: $KL[q_{\vparam}(\mparam)\|p(\mparam|\data)]$ (exclusive KL as used in standard VI) and  $KL[p(\mparam|\data)\|q_{\vparam}(\mparam)]$ (inclusive KL) respectively. There exist other equivalent $\alpha$-divergence definitions, e.g., Amari's \citep{amari1985differential}, which also include the two KL divergences as special cases.

Direct minimization of the $\alpha$-divergence in Eq.~\eqref{eq:alpha_divergence} is intractable again due to the intractable $p(\data)$. This is mitigated by again an equivalent maximization task on another lower bound of the log marginal likelihood, which can also be estimated with Monte Carlo, potentially with the reparameterization trick as well \citep{li2016renyi}:
\begin{equation}
     \begin{split}
        \mathcal{L}_{\alpha}(\vparam) &:=\log p(\data) - D_{\alpha}[ p(\mparam|\data) \| q_{\vparam}(\mparam) ]\\
        &= \frac{1}{1-\alpha} \log E_{q_{\vparam}(\mparam)}[(\frac{p(\data|\mparam)p(\mparam)}{q_{\vparam}(\mparam)})^{1-\alpha}]\\
        &\approx \frac{1}{1-\alpha} \log \frac{1}{M} \sum_{m=1}^M  (\frac{p(\data|\mparam_m)p(\mparam_m)}{q_{\vparam}(\mparam_m)})^{1-\alpha}, \qquad \mparam_m \stackrel{\text{iid}}{\sim} q_{\vparam}(\mparam)
    \end{split}
\end{equation}
%
Figure \ref{fig: alphadiv} illustrates the coverage of mean-field approximations fitted with $\alpha$-divergences, where the target distribution is a correlated Gaussian. Indeed by decreasing $\alpha$, one can interpolate between mode-seeking and mass-covering behaviors for the approximate posterior. 

\begin{center}
\begin{minipage}{.4\textwidth}
    \centering
    \includegraphics[width=0.9\textwidth]{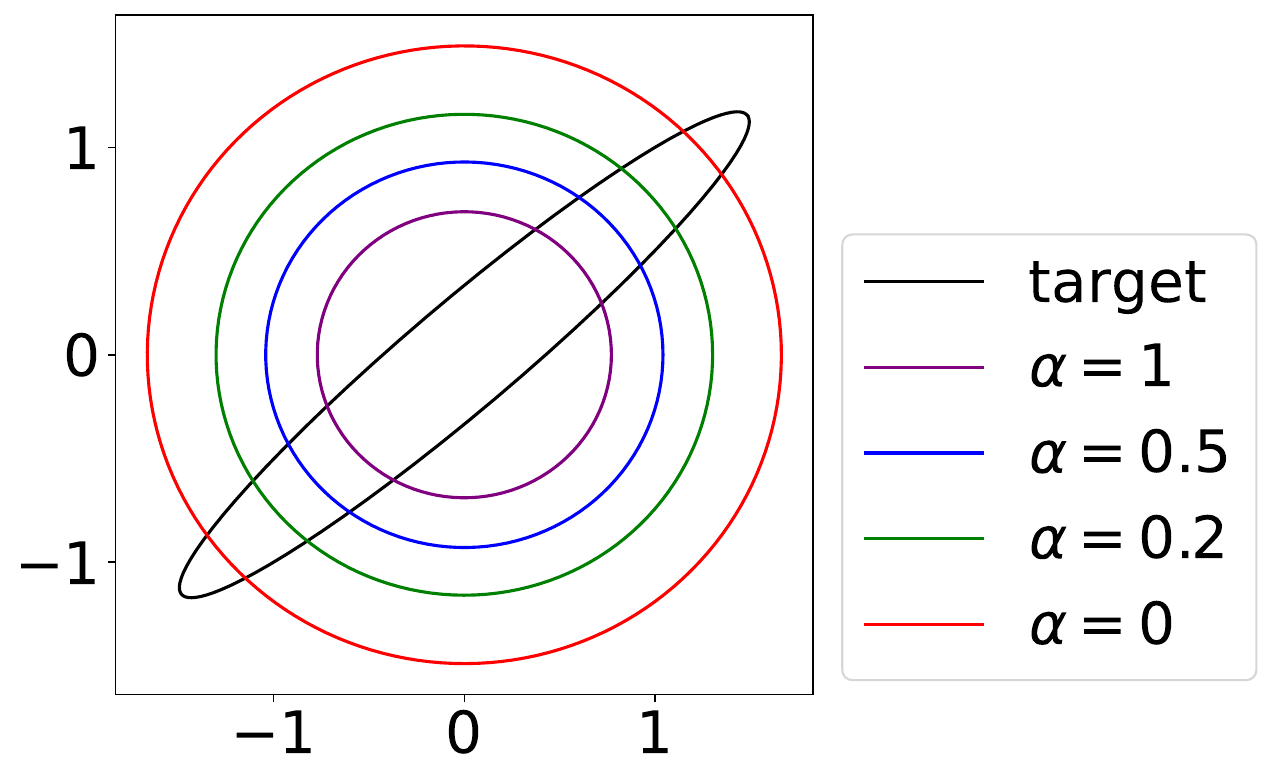}
\end{minipage}
\hfill
\begin{minipage}{.5\textwidth}
\captionof{figure}{Factorized Gaussians fitted by minimizing $\alpha$-divergences with different $\alpha$'s for a correlated 2D Gasussian target:}
\label{fig: alphadiv}
\begin{equation*}
\mathcal{N} \left(
\begin{pmatrix}
0\\
0
\end{pmatrix},
\begin{pmatrix}
2.0 & 1.5 \\
1.5 & 1.6 
\end{pmatrix} \right).
\end{equation*}
\end{minipage}
\end{center}

\subsubsection{Distribution Family}

One drawback of using mean-field Gaussian approximations is the negligence of covariance structures in the exact posterior. In this regard a Gaussian distribution with a full covariance matrix is favorable. However, storing and decomposing a covariance matrix require $O(d^2)$ memory and $O(d^3)$ time, respectively, which quickly become computationally intractable even if the number of BNN weights $d$ is around tens of thousands. Recall that modern neural networks for image recognition has millions, if not billions, of parameters \citep{krizhevsky2012imagenet,dehghani2023scaling}. Therefore Gaussians with freely parameterized full covariance matrices are not viable solutions, and further approximations are needed.

Recent attention on developing correlated Gaussian approximate posteriors focuses on so called ``low-rank $+$ diagonal'' structure for the covariance matrices. In other words, instead of searching over the entire space of positive semi-definite matrices for the covariance matrix, the Gaussian variational distribution $q_{\vparam}(\mparam)$ has the following structure \citep{tomczak2020efficient}:
\begin{equation}
    q_{\vparam}(\mparam) = \mathcal{N}(\vparam; \bm{\mu}, \mathbf{\Sigma}), \quad \mathbf{\Sigma} = \mathbf{L}\mathbf{L}^\top + \mathbf{D},
\end{equation}
where $\mathbf{L} \in \mathbb{R}^{d \times r}$ is a low-rank matrix (with $r << d$) and $\mathbf{D}$ is a diagonal matrix with positive diagonal entries. Here the variational parameters are $\vparam = \{\bm{\mu}, \mathbf{L}, \mathbf{D} \}$ which can be freely optimized using ELBO. In practice correlated Gaussian approximations are only applied layer-wise, i.e., the $q_{\vparam}$ distribution still factorizes over layers, but within each layer a low-rank Gaussian approximate posterior is fitted.

Another approach to introduce structured Gaussian approximations is to use matrix normal distributions. Notice that the popular choice of isotropic Gaussian prior $p(\weight^l) = \mathcal{N}(vec(\weight^l); \bm{0}, \sigma^2 \mathbf{I}_{d_o d_i})$ for the $l^{th}$ layer weight $\weight^l \in \mathbb{R}^{d_o \times d_i}$ has an equivalent matrix normal form $\mathcal{MN}(\weight^L; \bm{0}, \sigma_r^2\mathbf{I}_{d_o}, \sigma_c^2 \mathbf{I}_{d_i})$ with $\sigma_r > 0$ and $\sigma_c > 0$ the row and column standard deviations satisfying $\sigma = \sigma_r \sigma_c$. This inspires the design of matrix normal approximate posteriors $q_{\vparam}(\weight^l) = \mathcal{MN}(\weight^l; \mathbf{M}^l, \Sigma_r, \Sigma_c)$ with $\Sigma_r \in \mathbb{R}^{d_o \times d_o}$ and $\Sigma_c \in \mathbb{R}^{d_i \times d_i}$, reducing the costs to $\mathcal{O}(d_i^2 + d_o^2)$ memory and $\mathcal{O}(d_i^3 + d_o^3)$ time instead of $\mathcal{O}(d_i^2 d_o^2)$ and $\mathcal{O}(d_i^3 d_o^3)$ respectively for using full covariance Gaussian approximations. Further low-rank approximation techniques can be applied to parameterize $\Sigma_r$ and $\Sigma_c$ to further reduce the complexity figures \citep{louizos2017multiplicative,ritter2021sparse}.

Non-Gaussian posterior approximations have also been explored. Hierarchical conjugate priors \citep{kessler2021ibpbnn} and sparsity-inducing priors \citep{ghosh2019model,bai2020efficient} can be used to induce desirable structural properties of the weights, and in such cases posterior approximations typically share similar conjugacy or sparsity structure assumptions. 
Another line of work considers DNNs as flexible transformations to obtain expressive posterior approximations \citep{louizos2017multiplicative,tran:implicit2017,mescheder2017adversarial,li2018gradient}. In detail, with a learnable neural network $F_{\vparam}(\bm{\omega
})$ parameterized by $\vparam$ and operated on a potentially lower-dimensional variable $\bm{\omega}$, the approximate posterior $q_{\vparam} := (F_{\vparam})_{\#}q_0$ is defined as the push-forward distribution of a based distribution $q_0(\bm{\omega})$ that is often a standard Gaussian. 
When $F_{\vparam}(\bm{\omega
})$ represents an invertible function (e.g., normalizing flows \citep{rezende2015variational,louizos2017multiplicative, papamakarios2021normalizing}), the resulting approximate posterior density can be obtained via change-of-variable rule: $q_{\vparam}(\mparam) = q_0(F_{\vparam}^{-1}(\mparam)) | \frac{d F^{-1}_{\vparam}(\mparam)}{d \mparam}|$,
which can then be plugged into the ELBO (Eq.~\eqref{eq:elbo}), and techniques such as Monte Carlo estimate and the reparameterization trick are applicable.
On the other hand, for non-invertible $F_{\vparam}(\bm{\omega
})$ transforms, the resulting approximate posterior $q_{\vparam}(\mparam)$, while still supporting posterior predictive inference via Monte Carlo estimation and the reparameterization trick, no longer has a tractable density. Therefore further approximations to the ELBO objective and its gradients have been proposed \citep{mescheder2017adversarial,li2018gradient}. 
When $\bm{\omega}$ is lower-dimensional, these neural network transformed posterior may have mismatched support as compared with the prior. A solution to this issue is to add in e.g., Gaussian noise after neural network transform, resulting in an approximate posterior $q_{\vparam}(\mparam) = \int \mathcal{N}(\mparam | F_{\vparam}(\bm{\omega}), \sigma^2 \mathbf{I})q_0(\bm{\omega}) d\bm{\omega}$ which is then fitted using a lower-bound approximation to the ELBO \citep{salimans2015markov,yin2018semi}.

\vspace{-2em}
\section{Deep Generative Models}
\label{sec:dgm}
Deep generative models (DGMs) are a class of deep learning models that aim to learn a $\mparam$-parameterized distribution $p_{\mparam}(\bm{x})$ from the training set $\data=\{\bm{x}_i\}_{i=1}^N$ as an approximation for the true data distribution $p(\bm{x})$. After learning, these models can generate new data samples from $p_{\mparam}(\bm{x})$. DGMs have gained significant attention due to their ability to generate diverse and high-quality data, such as images~\citep{lugmayr2022repaint,saharia2022palette,kawar2023imagic}, text~\citep{gu2022vector,gong2022diffuseq,yang2023diffsound}, audio~\citep{chung2015recurrent,kong2020diffwave,huang2022fastdiff} and molecules~\citep{li2014generalized,collins2015energy,xu2021geodiff}.

Roughly speaking, DGMs can be categorized into two main types. The first type directly parameterizes the data distribution $p_{\bm{\theta}}(\bm{x})$. Examples include energy-based models~\citep{lecun2006tutorial} and score-based models~\citep{song2020score}. The second type introduces latent variables $\bm{z}$ to aid in the learning process and acquire latent representations. These models represent the data distribution as $p_{\bm{\theta}}(\bm{x})=\int p_{\bm{\theta}}(\bm{x}|\bm{z})p(\bm{z})d\bm{x}$. Examples include variational autoencoders~\citep{kingma2014auto}.

\subsection{Energy-based Models}
\label{sec:dgm_ebm}
Energy-based models (EBMs)~\citep{lecun2006tutorial} are probabilistic models with minimum restrictions on the functional form. They define an \emph{energy function} $E_{\mparam}(\bm{x})$ in the form of unnormalized log-probabilities. Practically, the energy function can be directly parameterized by any deep neural networks, which makes it flexible towards a variety of model architectures. The density given by EBMs is defined as:
\begin{equation}
    p_{\mparam}(\bm{x}):=\frac{1}{Z_{\mparam}}\exp{\left(-E_{\mparam}(\bm{x})\right)},
\end{equation}
where $Z_{\mparam}=\int_{\bm{x}}\exp{\left(-E_{\mparam}(\bm{x})\right)}d\bm{x}$ is the normalizing constant. This normalizing constant is a function of the model parameter $\mparam$, and thus cannot be ignored when learning $\mparam$.

\begin{figure}[t]
    \centering
    \includegraphics[width=\linewidth]{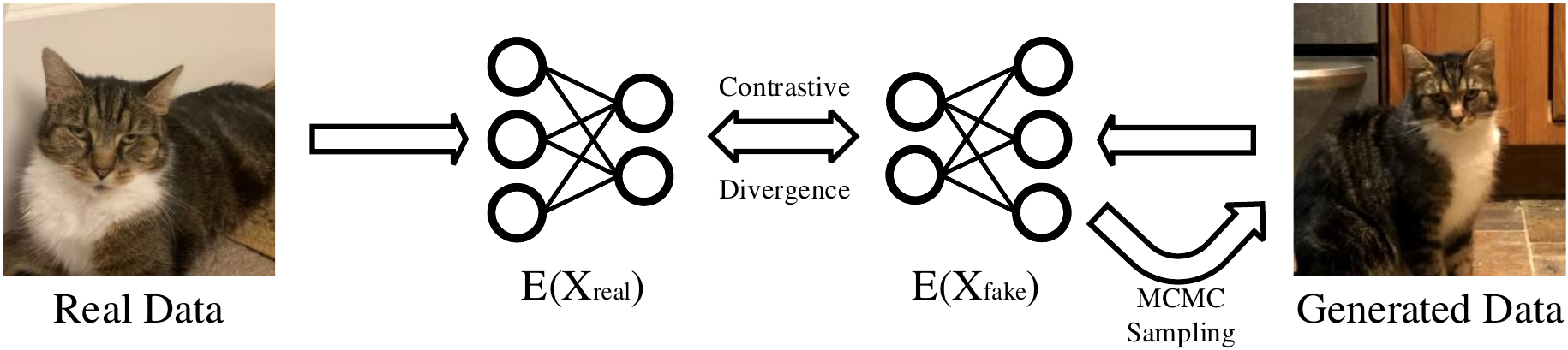}
    \caption{The training procedure of energy-based models. Generated data is obtained by sampling from the neural network-parameterized distribution $\bm{X}_{\text{fake}}\sim\exp(-E_{\mparam}(\bm{x}))$. The model parameter $\mparam$ is updated by minimizing the contrastive divergence between the real and generated data.}
    \label{fig:ebms}
\end{figure}

The ultimate goal of EBMs is to approximate the real data distribution $p(\bm{x})$ with the $\mparam$-parameterized density $p_{\mparam}(\bm{x})$. One standard way to achieve such an approximation is by the maximum likelihood estimation (MLE), in which the expected log-likelihood is maximized:
\begin{equation}
    \mparam^* := \max_{\mparam}E_{\bm{x}\sim p(\bm{x})} [\log p_{\mparam}(\bm{x})].
\end{equation}
The MLE approximation is equivalent to minimizing the KL-divergence $KL[p(\bm{x}) \| p_{\mparam}(\bm{x})]$. 
Unfortunately, gradient-based optimization cannot be directly applied to the MLE objective, since the the gradients of the logarithm of $p_{\mparam}(\bm{x})$ contains an intractable term $\nabla_{\mparam}\log Z_{\mparam}$: 
\begin{equation}
    \nabla_{\mparam}\log p_{\mparam}(\bm{x})=-\nabla_{\mparam}E_{\mparam}(\bm{x})-\nabla_{\mparam}\log Z_{\mparam}.
\label{eq:ebm_mle_gradient}
\end{equation}
A popular solution is based on Monte Carlo estimation with a random sample drawn from the EBM distribution, resulting in an approximate gradient to Eq.~\eqref{eq:ebm_mle_gradient}:
\begin{equation}
    \nabla_{\mparam}\log Z_{\mparam}=-E_{\bm{x}\sim p_{\mparam}(\bm{x})}[\nabla_{\mparam}E_{\mparam}(\bm{x})] \approx-\nabla_{\mparam}E_{\mparam}(\widetilde{\bm{x}}), \quad \widetilde{\bm{x}}\sim p_{\mparam}(\bm{x}),
\label{eq:ebm_mle_approx_gradient}
\end{equation}
\begin{equation}
    \Rightarrow \quad \nabla_{\mparam}\log p_{\mparam}(\bm{x}) \approx \nabla_{\mparam}E_{\mparam}(\widetilde{\bm{x}}) - \nabla_{\mparam}E_{\mparam}(\bm{x}), \quad \widetilde{\bm{x}}\sim p_{\mparam}(\bm{x}).
    \label{eq:grad_ebm}
\end{equation}
The training procedure of EBMs is illustrated in Fig.~\ref{fig:ebms}. As we shall see, sampling from the EBM is typically done via MCMC, often with finite number of transition steps. When using the empirical data distribution as the initial distribution, the corresponding approach is named \emph{contrastive divergence}~\citep{hinton2002training}, which is motivated as approximately minimizing the gap between two KL-divergences. 

\subsubsection{Langevin Sampling}
During both training and generation, we need to obtain samples $\widetilde{\bm{x}}$ from the EBM distribution $p_{\mparam}(\bm{x})$. However, $p_{\mparam}(\bm{x})$  lacks a closed form due to the intractable normalizing constant. Therefore, MCMC methods, particularly gradient-based MCMC, are commonly applied to acquire samples from $p_{\mparam}(\bm{x})$. For example, Langevin dynamics~\citep{parisi1981correlation,grenander1994representations,welling2011bayesian} utilizes the gradient of $\log p_{\mparam}(\bm{x})$:
\begin{equation}
    \nabla_{\bm{x}}\log p_{\mparam}(\bm{x})=-\nabla_{\bm{x}}E_{\mparam}(\bm{x})-\nabla_{\bm{x}}\log Z_{\mparam}=-\nabla_{\bm{x}}E_{\mparam}(\bm{x})
    \label{eq:gradient_log_p}
\end{equation}
The second equation holds because the normalizing constant $Z_{\mparam}$ is independent of $x$. Consequently,  the update rule for $x$ using Langevin dynamics is:
\begin{equation}\label{eq:langevin}
    \bm{x}_{k+1} \gets \bm{x}_k-\alpha_k\nabla_{\bm{x}_k}E_{\mparam}(\bm{x}_k)+\sqrt{2\alpha_k}\cdot\bm{\epsilon}_k,
\end{equation}
where $\bm{\epsilon}_k\sim\mathcal{N}(\bm{0},\bm{I})$ is a standard Gaussian noise and $\alpha_k$ is the stepsize. With an annealed step size schedule, $\bm{x}_k$ is guaranteed to converge to the EBM distribution $p_{\mparam}(\bm{x})$ as $k\rightarrow\infty$. 
During the training process, we can use the empirical data distribution as the initial distribution \citep{hinton2002training}, and run a short Markov chain to generate samples $\widetilde{\bm{x}}$ for computing the gradient in Eq.~\eqref{eq:grad_ebm}. The initialization of the MCMC can also be either random noise~\citep{nijkamp2019learning} or persistent states obtained from previous iterations~\citep{tieleman2008training}. A summary of the full training procedure for EBMs with MCMC sampling assistance is provided in Algorithm \ref{alg:ebm_training}. 


\begin{algorithm}[t]
\caption{Training Energy-based methods}\label{alg:ebm_training}
\begin{algorithmic}[1]
\Require Dataset $\data = \{ \x_n \}_{n=1}^N$, sampling step sizes $\{\alpha_k \}$, total optimization steps $T$, total sampling steps $K$.
\For{$t = 1, ..., T$}

\vspace{1.5mm}
\State $\triangleright$ Generate a sample from the current EBM via Langevin Sampling
\State Initialize $\bm{x}_0$ (see discussions in the main text)
\For{$t = 1, ..., K$}
\State $\bm{x}_{k+1} \gets \bm{x}_k-\alpha_k\nabla_{\bm{x}_k}E_{\mparam}(\bm{x}_k)+\sqrt{2\alpha_k}\cdot\bm{\epsilon}_k, ~~~\bm{\epsilon}_k\sim\mathcal{N}(\bm{0},\bm{I})$
\EndFor
\State Set the negative sample to be $\bm{x}^{-}=\bm{x}_K$
\vspace{1.5mm}
\State $\triangleright$ Maximize the log-likelihood of the EBM
\State Sample a datapoint from the dataset: $\bm{x}^{+}\sim \data$
\State Obtain the gradient of the log-likelihood: $\nabla_{\bm{x}}\log p_{\mparam}(\bm{x}) 
    = \nabla_{\mparam}  E_{\mparam} (\bm{x}^{-}) - \nabla_{\mparam}E_{\mparam} (\bm{x}^{+})$
\State Update $\mparam$ by MLE using gradient-based optimizers
\EndFor
\end{algorithmic}
\end{algorithm}

\subsubsection{Score Matching}
In Eq.~\eqref{eq:langevin}, generating new data only needs the gradient of the energy $\nabla_{\bm{x}_k}E_{\mparam}(\bm{x}_k)$ rather than the energy itself $E_{\mparam}(\bm{x}_k)$. Therefore, another way to learn EBMs is by directly parameterizing the gradient of $\log p_{\mparam}(\bm{x})$, referred to as the \emph{score function}: $s_{\mparam}(\bm{x})=\nabla_{\bm{x}}\log p_{\mparam}(\bm{x})\approx\nabla_{\bm{x}}\log p(\bm{x})$. Then the sampling rule during generation becomes:
\begin{equation}
    \bm{x}_{k+1} \gets \bm{x}_k+\alpha_ks_{\mparam}(\bm{x}_k)+\sqrt{2\alpha_k}\cdot\bm{\epsilon}_k.
\label{eq:langevin_score_matching}
\end{equation}
Generative models that learn the data distribution through the score function are called \emph{score-based models}~\citep{song2019generative}. One benefit of learning the score function is that it is free from the normalizing constant. From Eq.~\eqref{eq:gradient_log_p}, we have:
\begin{equation}
    s_{\mparam}(\bm{x})=\nabla_{\bm{x}}\log p_{\mparam}(\bm{x})=-\nabla_{\bm{x}}E_{\mparam}(\bm{x}).
\end{equation}
The training of score-based models involves \emph{score matching}~\citep{hyvarinen2005estimation}, in which the Fisher divergence between the learned score and true score is minimized:
\begin{equation}
    D_F(p(x) \| p_{\theta}(x)) = \frac{1}{2}E_{\bm{x}\sim p(\bm{x})}\left[\|s_{\mparam}(\bm{x})-\nabla_{\bm{x}}\log p(\bm{x})\|^2\right]\propto\mathbbm{E}_{\bm{x}\sim p(\bm{x})}\left[tr(\nabla_{\bm{x}}s_{\mparam}(\bm{x}))+\frac{1}{2}\|s_{\mparam}(\bm{x})\|^2\right].
\end{equation}
However, computing the trace of the Jacobian $tr(\nabla_{\bm{x}}s_{\mparam}(\bm{x}))$ is prohibitively costly for deep neural networks. To address this issue, \emph{denoising score matching} is proposed~\citep{vincent2011connection}, which introduces a small Gaussian perturbation to the original data $q(\tilde{\bm{x}}|\bm{x}) = N(\tilde{\bm{x}}|\bm{x}, \sigma^2I)$, and returns an accurate estimate of the score function when $\sigma \rightarrow 0$. Consequently,
\begin{align}
D_F(q(\tilde{\bm{x}}|\bm{x}) \| p_{\theta}(\tilde{\bm{x}})) 
&= \frac{1}{2} E_{p(\bm{x})}E_{q(\tilde{\bm{x}} | \bm{x})}[\| s_{\theta}(\tilde{\bm{x}}) - \triangledown_{\tilde{\bm{x}}} \log q(\tilde{\bm{x}} | \bm{x})\|^2]\nonumber \\
&= \frac{1}{2} E_{p(\bm{x})}E_{q(\tilde{\bm{x}} | \bm{x})}\left[\left\| s_{\theta}(\tilde{\bm{x}}) + \frac{\tilde{\bm{x}} - \bm{x}}{\sigma^2}\right\|^2\right]\nonumber\\
&=\frac{1}{2} E_{p(\bm{x})}E_{\epsilon\sim\mathcal{N}(0,I)}\left[\left\| s_{\theta}(\bm{x}+\sigma\epsilon) + \frac{\epsilon}{\sigma}\right\|^2\right]\label{eq:fisher-divergence}
\end{align}
In practice with $\sigma > 0$, the learned score approximates the score of the perturbed data distribution, not the original data distribution. Even if an accurate data score function estimate is obtained with $\sigma \approx 0$, sampling from the data distribution with Langevin dynamics (Eq.~\eqref{eq:langevin_score_matching}) remains challenging in e.g., fast mixing. To resolve these issues, \citet{song2019generative} proposes learning the score function across different noise scales via \emph{Noise Conditional Score Networks}, which parameterize the score function as $s_{\theta}(\bm{x},\sigma)$ with the noise scale $\sigma$ as an additional input. During generation, 
\emph{Annealed Langevin dynamics}~\citep{song2019generative} is used which simulates Eq.~\eqref{eq:langevin_score_matching} using $s_{\theta}(\bm{x},\sigma)$ with different $\sigma$, starting from the most noise-perturbed distribution (with $\sigma >> 0$) and then progressively reducing the noise scales until $\sigma \approx 0$. Consequently, the samples become close to the original data distribution.

\subsection{Diffusion Models}
\label{sec:dgm_ddpm}
\begin{figure}[t]
    \centering
    \includegraphics[width=\linewidth]{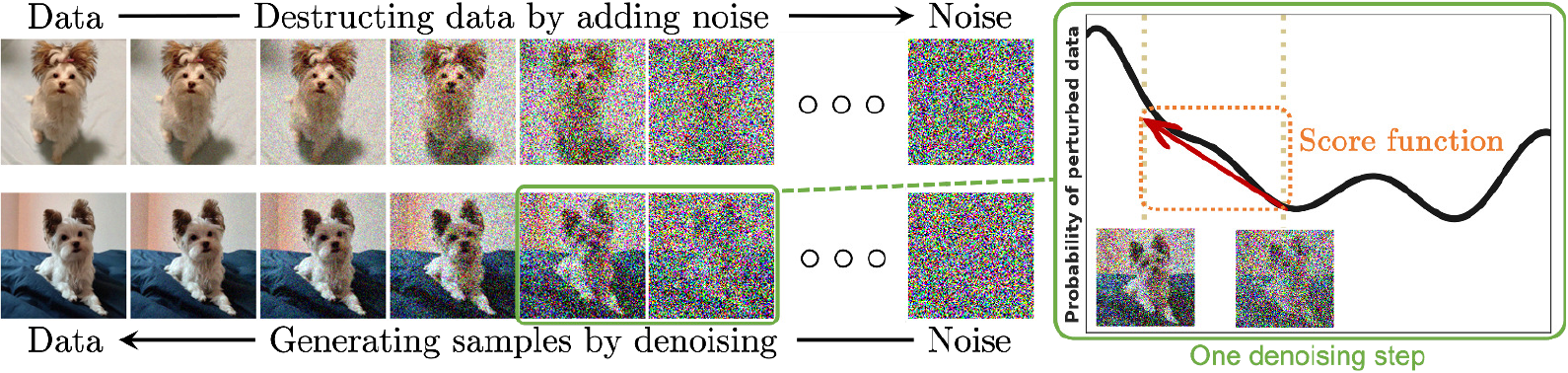}
    \caption{In diffusion models, the forward process progressively introduces noise, whereas the reverse process aims to denoise the perturbed data. The denoising step typically involves the estimation of the score function. Adapted from
\citet{yang2022diffusion}. Used with kind permission of Ling Yang.}
    \label{fig:diffusion}
\end{figure}

Diffusion models are probabilistic generative methods that generate new data by reversing the process of injecting noise into data~\citep{sohl2015deep,ho2020denoising}.
The intuition of diffusion models is illustrated in Fig.~\ref{fig:diffusion}.
Diffusion models are closely related to score-based models, essentially sharing the same training objective outlined in Eq.~\eqref{eq:fisher-divergence}, differing only by a constant. Subsequently, a continuous-time generalization of diffusion models through Stochastic Differential Equations (SDE) has been developed~\citep{song2020score}. The SDE perspective sheds light on the connection between diffusion models and MCMC methods, which we will discuss in detail below.   

\subsubsection{Stochastic Differential Equations}
The diffusion process that injects noise to data is governed by the following stochastic differential equation:
\begin{equation}
    d\bm{x}=\bm{f}(\bm{x},t)dt+g(t)d\bm{w},
\end{equation}
where $\bm{f}(\bm{x},t)$ and $g(t)$ are the diffusion and drift functions respectively, and $\bm{w}$ is a standard Wiener process (Brownian motion). \citet{anderson1982reverse} shows that any diffusion process can be reversed by solving the following reverse-time SDE:
\begin{equation}
    d\bm{x}=[\bm{f}(\bm{x},t)-g(t)^2\nabla_{\bm{x}}\log q_t(\bm{x})]dt+g(t)d\bar{\bm{w}},
    \label{eq:reverse_time_sde}
\end{equation}
where $\bar{\bm{w}}$ is a standard Wiener process when time flows backwards, and $dt$ denotes an infinitesimal negative time step. 

The reverse-time SDE can be simulated using various general-purpose numerical SDE solvers. To address the approximation errors arising from these numerical methods, gradient-based MCMC methods, specifically Langevin dynamics, have been employed.
The reason is that we
have the score function $s_{\theta}(\bm{x},t) \approx \nabla\log p_t(\bm{x})$ which can be used with gradient-based MCMC to sample from $p_t(\bm{x})$. This forms a hybrid sampler, denoted as \emph{Predictor-Corrector} (PC) sampler~\citep{song2020score}. Here, at each time step $t$, the numerical SDE solver first provides an estimate of the sample at the next time step, $\bm{x}_{t+1}$, serving as a ``predictor". Subsequently, the gradient-based MCMC method adjusts the marginal distribution of the estimated sample by running Monte Carlo sampling, using $\bm{x}_{t+1}$ as the initial value, serving as a ``corrector".

Recent developments of diffusion model's SDE design have been inspired by the continuous-time MCMC literature. Notably, continuous-time diffusion models have been found to align with overdamped Langevin dynamics with high friction coefficients~\citep{dockhorn2021score}. To enhance the convergence towards equilibrium, a critically-damped Langevin diffusion has been introduced, leveraging concepts from Hamiltonian dynamics~\citep{dockhorn2021score}.


\vspace{-1em}
\subsection{Deep Latent Variable Models}
\label{sec:dgm_lvm}
\begin{figure}[t]
    \centering
\includegraphics[width=0.3\textwidth]{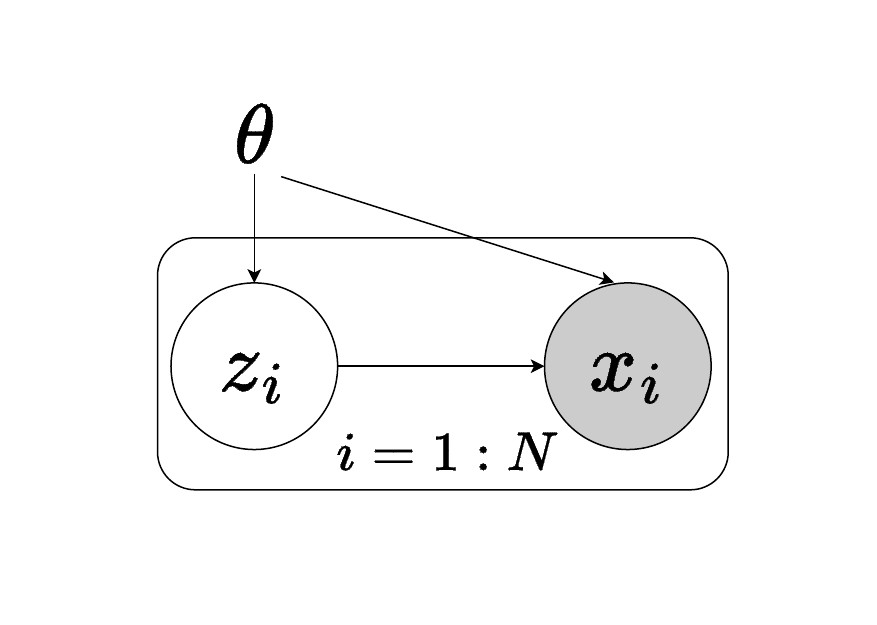}
    \vspace{-1.5em}
    \caption{Graphical model of a deep latent variable model.}
    \label{fig: dlvm}
\end{figure}

Apart from generating realistic data and accurate density estimation where EBMs/SBMs and diffusion models are well suited, another important application of deep generative models is representation learning, possibly with dimensionality reduction. In such scenario we are interested in obtaining a (possibly lower dimensional) representation $\latent$ for a given datapoint $\bm{x}$ sampled from the underlying data distribution. A starting example is Probabilistic PCA \citep{tipping1999probabilistic} which assumes the following model:
\begin{equation}
\latent \sim p_{\mparam}(\z) := \mathcal{N}(\bm{0}, \mathbf{I}), \quad \bm{x} \sim p_{\mparam}(\bm{x} | \latent) := \mathcal{N}(\weight \latent + \mathbf{b}, \sigma^2 \mathbf{I}).
\label{eq:probabilistic_pca}
\end{equation}
The model parameters $\mparam = \{ \weight, \mathbf{b}, \sigma \}$ is then fitted via maximum likelihood estimation (MLE) given a dataset $\data=\{\bm{x}_n\}_{n=1}^N$:
\begin{equation}
    \mparam^* = \arg\max_{\mparam} \sum_{n=1}^N \log p_{\mparam}(\bm{x}_n), \quad p_{\mparam}(\bm{x}_n) = \int p_{\mparam}(\bm{x} | \latent) p_{\mparam}(\latent) d\latent.
    \label{eq:dlvm_mle}
\end{equation}
Since both the prior and the likelihood are Gaussians, the log marginal likelihood $\log p_{\mparam}(\bm{x})$ is tractable, and there exist analytic solutions for the model parameters $\mparam$.

(Probabilistic) PCA returns meaningful representations when the underlying data generation process follows a linear transform of some latent features. Therefore it is less suited for typical deep learning applications with inputs as e.g., images, where useful features must be obtained in a non-linear way.
Instead, Deep Latent Variable Models (DLVMs) \citep{kingma2014auto,rezende:vae2014} are typically used to extract non-linear features from such high-dimensional and complex datapoints, since they are generative models that assume a data-generating process as transforming latent variables through a neural network. A typical DLVM assumes the observation $\bm{x}$ is generated from the conditional distribution $p_{\mparam}(\bm{x}|\latent)$ parameterized by a DNN with parameters $\mparam$. Compared with Probabilistic PCA, the conditional distribution in Eq.~\eqref{eq:probabilistic_pca} is now replaced by $p_{\mparam}(\bm{x} | \latent) = \mathcal{N}(\bm{x}; f_{\mparam}(\latent), \sigma^2 \mathbf{I})$ with a neural network $f_{\mparam}(\latent)$. It can then be viewed as a non-linear extension of Probabilistic PCA, since this formulation recovers Probabilistic PCA if $f_{\mparam}(\latent)$ is defined as a linear function.
The model parameters $\mparam$ in a DLVM is also estimated via maximum likelihood estimation, however, here the MLE objective (Eq.~\eqref{eq:dlvm_mle}) becomes intractable, due to the complex integral w.r.t.~the latent variables $\latent_n$ when a non-linear function $f_{\mparam}(\z)$ is in use. In practice, we again resort to approximation, where popular approaches are often VI-based \citep{kingma2014auto,rezende:vae2014}. Different from VI in BNNs, amortized inference \citep{gershman2014amortized} plays a key role in VI for DLVMs to further reduce computational costs. As we shall see, instead of constructing individual approximations to the posterior $p_{\mparam}(\latent_n|\bm{x}_n)$ for each $\bm{x}_n$, another 
DNN is employed to directly map $\bm{x}_i$ to the variational parameters of its corresponding variational distribution.

\subsubsection{Variational Autoencoders}
Variational Autoencoders (VAEs) \citep{kingma2014auto,rezende:vae2014} are a class of DLVMs paired with \emph{inference networks} (also called encoder networks) for amortized VI. Typically the Gaussian variational distribution family is used: $q_{\vparam}(\latent|\bm{x})=\mathcal{N}(\bm{\mu}_{\vparam}(\bm{x}), \bm{\sigma}^2_{\vparam}(\bm{x}))$, where the posterior mean and variance for an observation $\bm{x}$ are obtained by an inference network $g_{\vparam}(\bm{x}) := [\bm{\mu}_{\vparam}(\bm{x}), \log \bm{\sigma}_{\vparam}(\bm{x})]$ with variational parameters $\vparam$.
The model parameters $\mparam$ and the variational parameters $\vparam$ can be jointly trained by maximizing the ELBO: 
\begin{equation}
\begin{aligned}
\mparam^*, \vparam^* &= \arg\max_{\mparam, \vparam} \sum_{n=1}^N \mathcal{L}_{ELBO}(\bm{x}_n, \mparam, q_{\vparam}), \\
\log p_{\mparam}(\bm{x}) \geq \mathcal{L}_{ELBO}(\bm{x}, \mparam, q_{\vparam}) &= E_{q_{\vparam}(\latent|\bm{x})}[\log p_{\mparam}(\bm{x}|\latent)] - KL[q_{\vparam}(\latent|\bm{x})\|p_{\mparam}(\latent)].
\end{aligned}
\label{eq:vae_objective}
\end{equation}
Again the intractable expectation can be approximated with Monte Carlo estimation (often with only 1 sample), and the reparameterization trick \citep{kingma2014auto,rezende:vae2014} is also applied to enable back-propagation:
\begin{equation}
    \mathcal{L}_{ELBO}(\bm{x}, \mparam, q_{\vparam}) \approx \log p_{\mparam}(\bm{x}|\bm{\mu}_{\vparam}(\bm{x})+ \bm{\sigma}_{\vparam}(\bm{x}) \odot \bm{\epsilon}) - KL[q_{\vparam}\|p_{\mparam}], \quad \bm{\epsilon} \sim \mathcal{N}(\bm{0}, \mathbf{I}).
\label{eq:vae_objective_mc_reparam}
\end{equation}
This ELBO optimization can be viewed as training a stochastic autoencoder (a stochastic encoder $q_{\vparam}(\latent | \bm{x})$ plus a stochastic decoder $p_{\mparam}(\bm{x} | \latent)$), hence the name ``autoencoder''. The full algorithm for training VAEs is provided as Algorithm \ref{alg:vae} with stochastic gradient descent.

\begin{algorithm}[t]
\caption{SGD Training for a variational auto-encoder with Gaussian $q$ distribution}\label{alg:vae}
\begin{algorithmic}[1]
\Require Dataset $\data = \{ \x_n \}_{n=1}^N$, initialized parameters $\mparam, \vparam$,  batch-size $B$, SGD step sizes $\{\alpha_t \}$, SGD total steps $T$.
\For{$t = 1, ..., T$}
\State Sample a mini-batch of datapoints $\bm{\Xi} = \{ \x \} \sim \data^B$ with $| \bm{\Xi} | = B$
\For{$\x \in \bm{\Xi}$}
    \State Encode the input data: $g_{\vparam}(\bm{x}) \gets [\bm{\mu}_{\vparam}(\bm{x}), \log \bm{\sigma}_{\vparam}(\bm{x})]$
    \State Compute the KL term in the ELBO:
    $\mathcal{L}_{ELBO}(\x, \mparam, q_{\vparam}) \gets - KL[q_{\vparam}(\z | \x)\|p(\z)]$
    \State Sample the latent variable: $\z \gets \bm{\mu}_{\vparam}(\bm{x}) + \bm{\sigma}_{\vparam}(\bm{x}) \odot \bm{\epsilon}$, $\bm{\epsilon} \sim \mathcal{N}(\bm{0}, \mathbf{I})$
    \State Decode the latent variable to $f_{\mparam}(\z)$
    \State Add-in data likelihood: $\mathcal{L}_{ELBO}(\x, \mparam, q_{\vparam}) \gets \mathcal{L}_{ELBO}(\x, \mparam, q_{\vparam}) + \log p(\x| \z)$
\EndFor
\State Compute SGD updates:
$(\mparam, \vparam) \gets (\mparam, \vparam) + \alpha_t \nabla_{(\mparam, \vparam)} \frac{1}{|\bm{\Xi}|} \sum_{\x \in \bm{\Xi}} \mathcal{L}_{ELBO}(\x, \mparam, q_{\vparam})$
\EndFor
\end{algorithmic}
\end{algorithm}

The class of VAE models can be extended by considering VI with alternative divergences. For example, we can again control the coverage behavior of $q_{\vparam}(\latent|\bm{x})$ (from mode-seeking to mass covering) by using $\alpha$-divergences with different $\alpha$ hyper-parameters \citep{miguel2015alpha,li2016renyi}. The standard VAE can be viewed as a special case within this framework where the exclusive KL-divergence ($\mathrm{KL}[q\|p]$) is chosen within the $\alpha$-divergence family. Another special case is the use of the inclusive KL-divergence ($\mathrm{KL}[p\|q]$), the resulting model becomes an Importance-Weighed Autoencoder (IWAE) \citep{burda2016iwae}, which was originally proposed to tighten the lower bound of $\log p_{\mparam}(\bm{x})$ for optimization:
\begin{equation}
\hspace{-0.7em}
\log p_{\mparam}(\bm{x}) \geq \mathcal{L}_{IWAE}(\bm{x}, \mparam, q_{\vparam})
:= E_{\prod_{m=1}^M q_{\vparam}(\latent^{(m)}|\bm{x})}\left[\log \frac{1}{M} \sum_{m=1}^M \left[\frac{ p_{\mparam}(\bm{x}|\latent^{(m)})p_{\mparam}(\latent^{(m)})}{q_{\vparam}(\latent^{(m)}|\bm{x})} \right] \right]. 
\end{equation}
Again Monte Carlo estimation and the reparameterization trick apply to IWAE. The lower bound can be made tighter by using more MC samples, and as $M \rightarrow \infty$, $\mathcal{L}_{IWAE}$ converges to $\log p_{\mparam}(\bm{x})$. While benefiting learning of model parameters $\mparam$, in such case the fitting of $q_{\vparam}(\latent | \bm{x})$ may be sub-optimal, as any reasonable importance sampling proposal can return the exact marginal likelihood with infinite amount of samples \citep{rainforth2018tighter}.

\subsubsection{Combining VI and MCMC in DLVM}
The ELBO (Eq.~\eqref{eq:vae_objective}) is a biased approximation to the MLE objective (Eq.~\eqref{eq:dlvm_mle}) unless the variational posterior $q_{\vparam}(\latent_i | \bm{x}_i)$ matches the exact posterior $p_{\mparam}(\latent_i | \bm{x}_i)$. This motivates the use of MCMC to generate more accurate approximate posterior samples whose distribution asymptotically converges to the exact posterior \citep{doucet2023differentiable,hoffman2017learning}. Starting from an initial distribution $q_0(\latent | \bm{x})$, one runs $T$ steps of MCMC transitions to obtain an improved approximate posterior $q_T(\latent | \bm{x})$, and optimizes the model parameters $\mparam$ by maximizing $\mathcal{L}_{ELBO}(\bm{x}, \mparam, q_T)$ with the MCMC sample distribution $q_T(\latent | \bm{x})$ as the approximate posterior.
In DLVM context, gradient-based samplers such as Hamiltonian Monte-Carlo (HMC) \citep{neal2010mcmc,duane1987hybrid} are preferred, since they adapt to the decoder more efficiently than e.g., Metropolis-Hastings \citep{metropolis1953equation} with random walk proposals \citep{metropolis1953equation,sherlock2010random}. Still MCMC is computationally more expensive than VI and in practice may take prohibitively many transitions to converge, especially when the posterior is high-dimensional.

To speed-up MCMC, a popular approach is to draw its initial samples from a distribution $q_0$ that is already close to the exact posterior, with the hope of reducing the number of MCMC transition steps to obtain high-quality posterior samples \citep{hoffman2017learning,geffner2021mcmc}. A natural choice for such initial distributions is the variational posterior: $q_0^{(\vparam)}(\latent|\bm{x})=q_{\vparam}(\latent|\bm{x})$, where one can fit the variational parameters $\vparam$ by maximizing $\mathcal{L}_{ELBO}(\bm{x}, \mparam, q_{\vparam})$. 
Another approach involves distilling the initial distribution using the feedback from the final samples of a MCMC: one can improve $q_0^{(\vparam)}(\latent|\bm{x})$ by minimizing  $D(q_0^{(\vparam)}(\latent|\bm{x}), q_T(\latent|\bm{x}))$ for a chosen distance/divergence measure between distributions. Note that $q_T(\latent|\bm{x})$ implicitly depends on $\vparam$ but in practical implementations we apply a ``stop gradient'' operation to it and treat it as a constant w.r.t.~$\vparam$. Also the density of $q_T(\latent|\bm{x})$ is unknown as it is implicitly defined based on MCMC samples. These suggest the use of the inclusive KL-divergence $KL[q_T(\latent|\bm{x})\|q_0^{(\vparam)}(\latent|\bm{x})] = E_{q_T}[-\log q_0^{(\vparam)}(\latent|\bm{x})] + C$ for such divergence measure, as minimizing it w.r.t.~to $\vparam$ doesn't require the density of $q_T(\latent|\bm{x})$ \citep{li2017mcmc}. 
\citet{pmlr-v97-ruiz19a} further proposed a new divergence tailored to this task.

\begin{figure}[t]
    \centering
\includegraphics[width=0.7\textwidth]{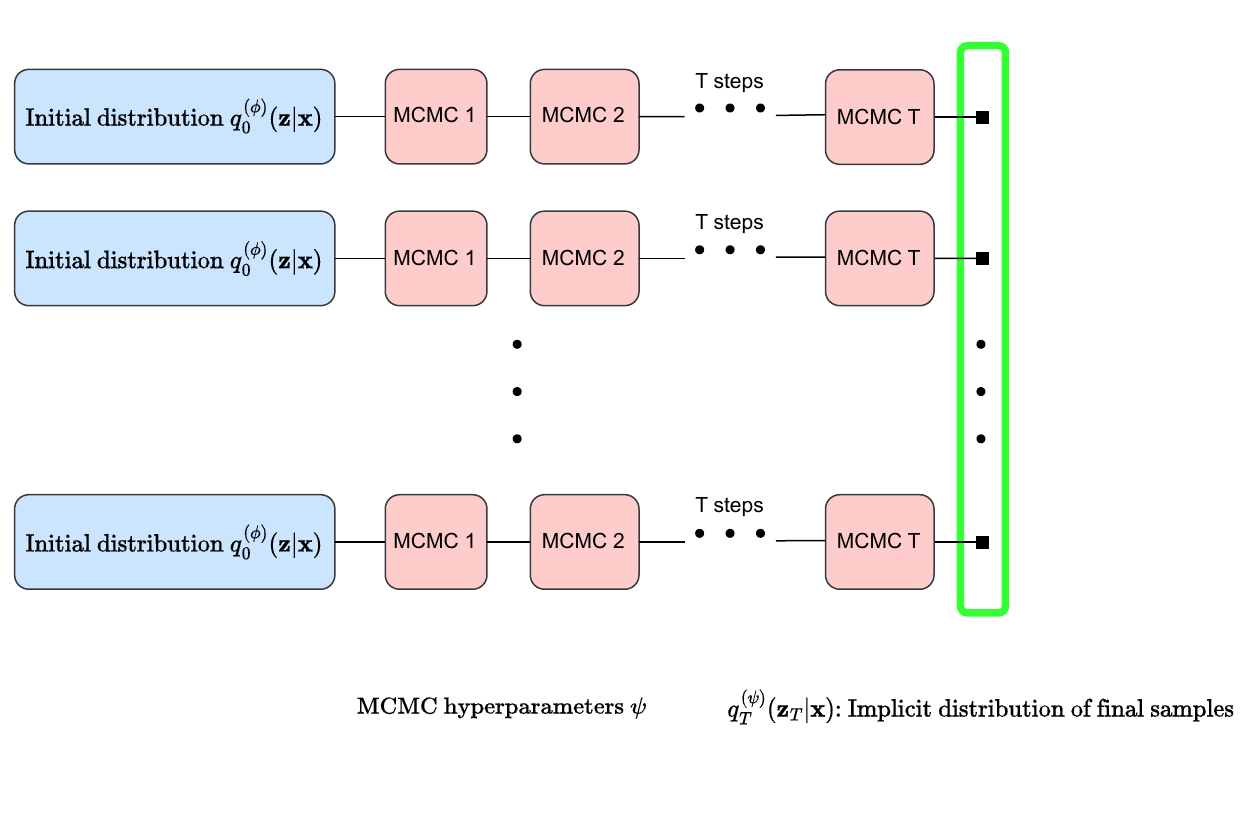}
\vspace{-2em}
    \caption{Implicit variational distribution $q_T^{(\psi)}(\latent_T|\bm{x})$ defined via a T-step MCMC. The hyper-parameters of MCMC $\psi$ are treated as learnable variational parameters.}
    \label{fig: mcmc_as_vi}
\end{figure}

In addition to improving the initial distributions, variational inference can also be applied to tuning hyper-parameters of MCMC samplers \citep{salimans2015markov,caterini2018hamiltonian,gong2019meta,campbell2021gradient}, such as the step size and momentum variance in HMC, which are known to be critical for fast convergence of the samplers \citep{neal2010mcmc}. A line of research, as illustrated in Figure \ref{fig: mcmc_as_vi}, treats the MCMC hyper-parameters $\psi$ as additional variational parameters, meaning the improved approximate posterior $q_T^{(\psi)}(\latent_T|\bm{x})$ can be further optimized w.r.t.~$\psi$ based on certain divergences between the exact posterior and the implicit distribution.
Since $q_T^{(\psi)}(\latent_T|\bm{x})$ is now implicit, its density is intractable, therefore the ELBO $\mathcal{L}_{ELBO}(\bm{x}, \mparam, q_T^{(\psi)})$ cannot be computed. In particular, the ELBO $\mathcal{L}_{ELBO}(\bm{x}, \mparam, q_T^{(\psi)})$ in Eq.~\eqref{eq:vae_objective} can also be reformulated as follows:
\begin{equation}
    \mathcal{L}_{ELBO}(\bm{x}, \mparam, q_T^{(\psi)}) = E_{q_T^{(\psi)}(\latent_T|\bm{x})}[\log p_{\mparam}(\bm{x}, \latent_T)] + E_{q_T^{(\psi)}(\latent|\bm{x})}[-\log q_T^{(\psi)}(\latent_T|\bm{x})].
\label{eq:vae_elbo_entropy}
\end{equation}

Maximizing the differential entropy $H[q_T^{(\psi)}]= E_{q_T^{(\psi)}(\latent|\bm{x})}[- \log q_T^{(\psi)}(\latent_T|\bm{x})]$ prevents $q_T^{(\psi)}$ from collapsing to a point mass located at a mode of the true posterior $p_{\theta}(\latent|\bm{x})$. To circumvent the intractability of $q_T^{(\psi)}(\latent|\bm{x})$ density, \citet{salimans2015markov} proposed a variational inference approach in the expanded space of all intermediate states along the Markov chain (i.e. $\latent_{0:T}$), and derived a lower bound of $\mathcal{L}_{ELBO}$ as the optimization objective without computing $H[q_T^{(\psi)}]$. 
Noticing that gradient-based optimization is in use, \citet{li2018gradient} and \citet{gong2019meta} proposed to directly approximate the gradient of $H[q_T^{(\psi)}]$ with respect to $\psi$ in the optimization procedure.
\citet{campbell2021gradient} proposed to simply drop $H[q_T^{(\psi)}]$ from the ELBO (Eq.~\eqref{eq:vae_elbo_entropy}) and prevent $q_T^{(\psi)}$ from collapsing by using a mass-covering initial distribution $q_0^{(\vparam)}$. 
Alternatively, one can also bypass the intractability of the density of $q_T^{(\psi)}$ by using divergences which only require samples from $q_T^{(\psi)}$ as objective, for example Kernelized Stein Discrepancy \citep{Liu2016kernel,chwialkowski2016kernel}. 

\section{Discussion}
We have discussed scalable MCMC and approximate inference techniques for Bayesian computation in deep learning, with applications to quantifying uncertainty in DNNs and training deep generative models. 
Advances in fast posterior inference have contributed significantly to the rapid development of Bayesian neural networks as well as the profound success of deep generative models. Bayesian computation has been, and will continue to be, instrumental for embedding probabilistic principles into deep learning models.

The emphases on the need of posterior inference exhibit differences in the discussed two categories of modeling applications. For BNNs, posterior inference is the main computation tool for predictions and uncertainty estimates, although approximate inference for BNNs may require fitting parametric approximate posteriors, thereby introducing the concept of ``BNN training'' similar to training deterministic DNNs \citep{blundell2015bbp,gal2016thesis,wilson2020bayesian}. For deep generative models, posterior inference serves as part of the training routine, and it may not be needed for data generation tasks in test time. Interestingly, when considering DLVMs for unsupervised or semi-supervised learning, posterior inference becomes handy again in obtaining useful representations of the observed data, which will then be used for further downstream tasks in test time \citep{kingma2014semi,higgins2016beta,tschannen2018recent}. An interesting recipe in this case is to use fast inference (e.g., mean-field VI) in DLVM training, and high-fidelity approaches (e.g., MCMC with rejection steps) in test time to obtain more accurate and robust inference results \citep{kuzina2022alleviating}.

We conclude this chapter by discussing a few important research challenges to be addressed for more accurate and scalable Bayesian computation in deep learning context. 

\vspace{-1em}
\begin{itemize}
    \item For many popular DNNs the network weights are non-identifiable due to weight symmetries \citep{chen1993geometry,phuong2020functional}. This leads to many symmetric modes in the exact posterior of a BNN and increased difficulties of MCMC sampling \citep{papamarkou2022challenges,izmailov2021bayesian}. But ultimately one cares more about the \emph{posterior predictive distribution} (Eq.~\ref{eq:bayespred}) of the neural network outputs, where averaging over symmetric modes in the exact posterior provides no additional advantage in this regard. So future research efforts should focus on accurate computation of the predictive posterior, which may provide exciting opportunities for fast and memory-efficient weight-space posterior approximations \citep{sun2019functional,ma2019variational,ritter2021sparse}. Optimization aspects are also under-explored for both SG-MCMC inference and parametric approximate posterior fitting of BNN posteriors. Apart from incorporating popular adaptive stochastic gradient methods \citep{li2016preconditioned,chen2016bridging}, the study of neural network loss landscape \citep{li2018visualizing,maddox2019simple} may also inspire novel algorithms and analyses of posterior inference methods. 

    \item For training deep generative models with posterior inference as a sub-routine, the bias in the training procedure (as compared with e.g., MLE), resulted from errors in inference, is much less well understood. For example, contrastive divergence \citep{hinton2002training} as an approximation to MLE for energy-based models has been shown to perform adversarial optimization \citep{yair2021contrastive}, but it is still unclear theoretically about the impact of MCMC convergence to the quality of learned energy-based model. For variational auto-encoders, although research has discussed the impact of amortization gap \citep{cremer2018inference,marino2018iterative} in variational inference and the aforementioned research in combining MCMC and VI, the combined impact of amortization and restricted variational family on latent variable model training is still an open question.

\end{itemize}

\subsubsection*{A note on author contributions}
Wenlong Chen and Yingzhen Li wrote the sections regarding variational inference training for both Bayesian neural networks and deep latent variable models. Bolian Li and Ruqi Zhang wrote the sections on MCMC methods for Bayesian neural networks, energy-based models and diffusion models. Ruqi Zhang and Yingzhen Li edited the chapter in its final form for consistent presentations.

\bibliographystyle{apalike} 
\bibliography{references}

\end{document}